\pdfoutput=1
\documentclass[11pt]{article}

\usepackage[final]{acl}
\usepackage{times}
\usepackage{latexsym}

\usepackage[T1]{fontenc}
\usepackage[utf8]{inputenc}

\usepackage{microtype}

\usepackage{inconsolata}

\usepackage{graphicx}

\usepackage{xcolor}
\usepackage{booktabs}
\usepackage{fancyvrb}
\usepackage{xcolor}
\usepackage{lscape}
\usepackage{tabularx}
\usepackage{enumitem}

\title{Reassessing Graph Linearization for Sequence-to-sequence AMR Parsing: On the Advantages and Limitations of Triple-Based Encoding}

\author{ \textbf{Jeongwoo Kang\textsuperscript{1}} \quad
 \textbf{Maximin Coavoux\textsuperscript{1}} \quad
 \textbf{Cédric Lopez\textsuperscript{2}} \quad
 \textbf{Didier Schwab\textsuperscript{1}} \\
 \textsuperscript{1}Univ. Grenoble Alpes, CNRS, Grenoble INP,  LIG, 38000 Grenoble, France \\
 \textsuperscript{2}Emvista, Immeuble Le 610, 10 Rue Louis Breguet Bâtiment D, 34830 Jacou, France \\ 
\textsuperscript{1}\nolinkurl{{firstname}.{lastname}@univ-grenoble-alpes.fr} \\
\textsuperscript{2}\nolinkurl{{firstname}.{lastname}@emvista.com}
}

\begin{document}
\maketitle
\begin{abstract}

Sequence-to-sequence models are widely used to train Abstract Meaning Representation \citep[AMR]{banarescu-etal-2013-abstract} parsers. To train such models, AMR graphs have to be linearized into a one-line text format. While Penman encoding is typically used for this purpose, we argue that it has limitations: (1) for deep graphs, some closely related nodes are located far apart in the linearized text (2) Penman's tree-based encoding necessitates inverse roles to handle node re-entrancy, doubling the number of relation types to predict. To address these issues, we propose a triple-based linearization method and compare its efficiency with Penman linearization. Although triples are well suited to represent a graph, our results suggest room for improvement in triple encoding to better compete with Penman’s concise and explicit representation of a nested graph structure.

\end{abstract}

\section{Introduction}\label{sec:introduction}

\begin{figure}
    \centering
    \includegraphics[width=0.75\linewidth]{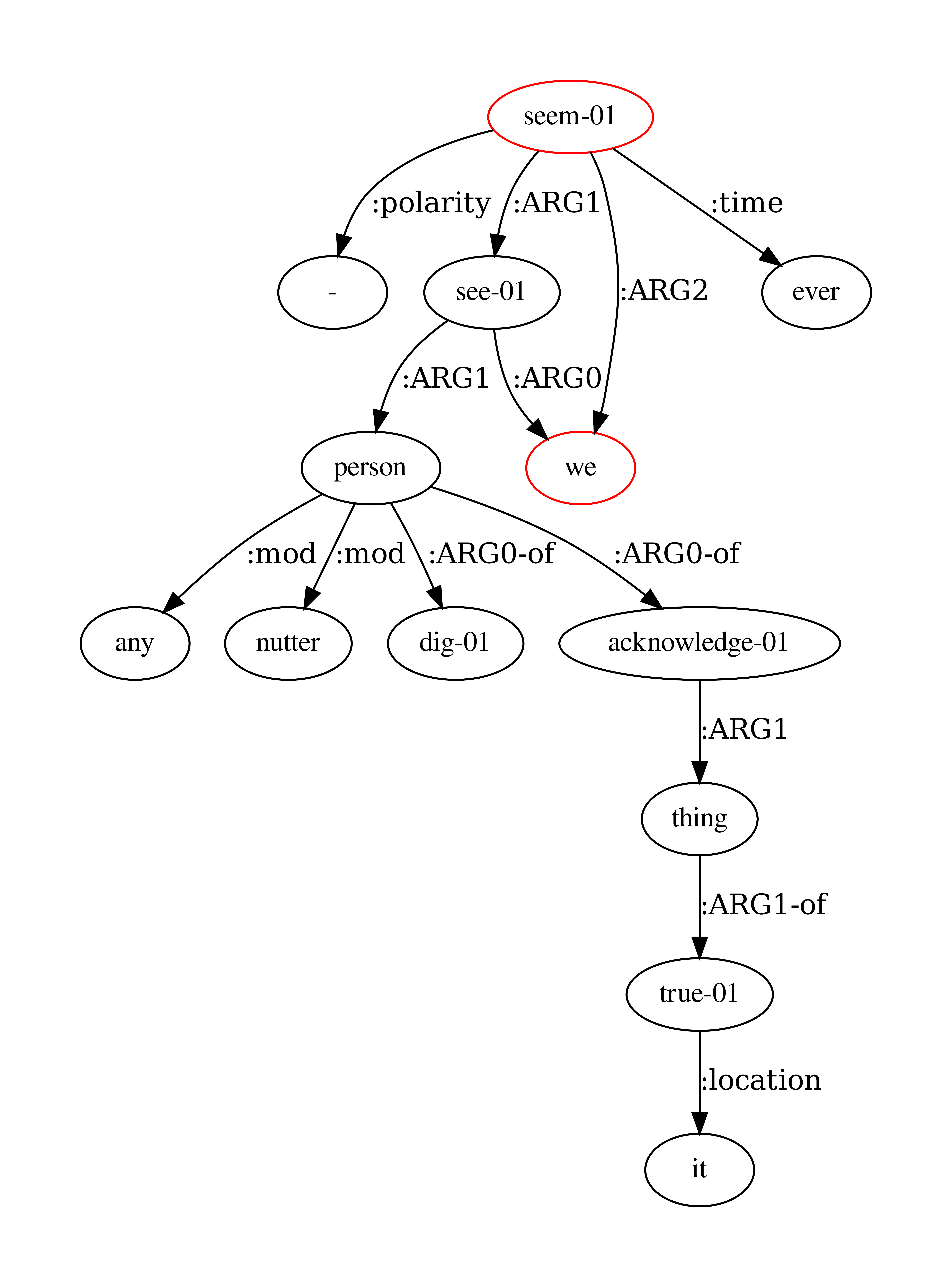}
    \caption{AMR graph for ``We never seem to see any of the dug-in nutters acknowledge the truth in it.'' 
    Example from the AMR~3.0 dataset \citep{knight-etal-2020-ldc2020}.}
    \label{fig:amr_graph_distance}

    \centering
    \begin{scriptsize}
    \begin{Verbatim}[commandchars=\\\{\}]
(s / \textbf{\textcolor{red}{seem-01}} :polarity -
        :ARG1 (s2 / see-01
            :ARG0 w
            :ARG1 (p / person
                    :mod (a / any)
                    :mod (n / nutter)
                    :\textbf{\textcolor{blue}{ARG0-of}} (d / dig-01)
                    :\textbf{\textcolor{blue}{ARG0-of}} (a2 / acknowledge-01
                        :ARG1 (t / thing
                                :\textbf{\textcolor{blue}{ARG1-of}} (t2 / true-01
                                    :location (i / it))))))
        :ARG2 (w / \textbf{\textcolor{red}{we}})
        :time (e / ever))
    \end{Verbatim}
    \end{scriptsize}
    \caption{AMR in Penman encoding for Figure \ref{fig:amr_graph_distance}.}
    \label{fig:amr_penman_distance}

    \centering
    \includegraphics[width=\linewidth]{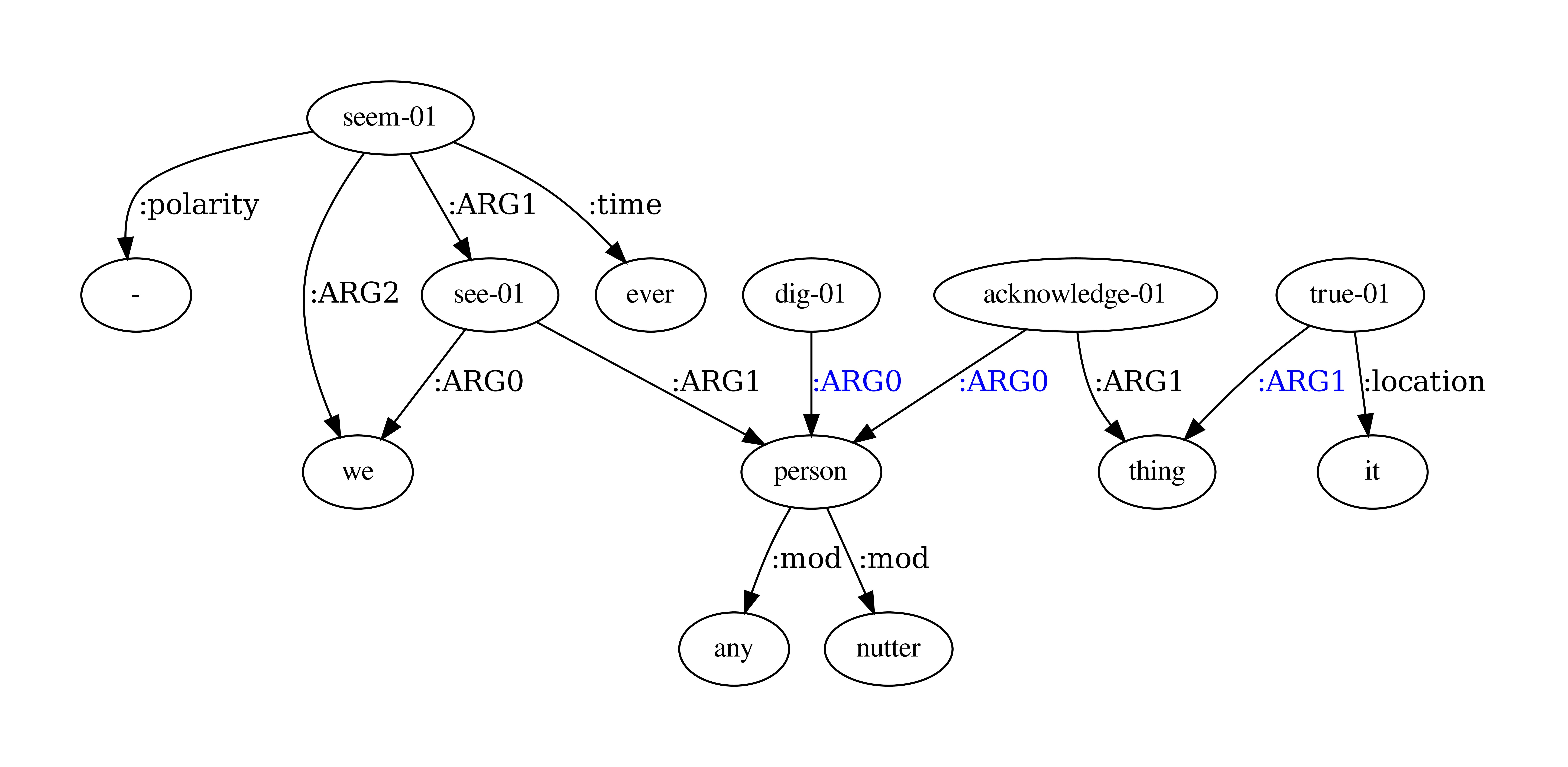}
    \caption{AMR graph without inverse roles.}
    \label{fig:amr_without_invroles}
\end{figure}

Abstract Meaning Representation (AMR) captures text meaning, such as "who does what to whom," and represents it in graphs (see Figure~\ref{fig:amr_graph_distance}). Structured information is easier for computers to process and therefore, AMR is widely used in NLP applications, e.g., machine translation \citep{wein-schneider-2024-lost}, text generation \citep{huang-etal-2023-paraamr}, or human-robot interaction systems \citep{bonial-etal-2019-abstract,bonial-etal-2023-abstract}.  

Sequence-to-sequence (seq2seq) approaches have recently gained popularity for AMR parsing due to strong performance and easy implementation. For prediction, the model receives an input sentence and outputs an AMR graph in text format. To train seq2seq models for AMR parsing, graph linearization to represent an AMR graph in a one-line text format is a prerequisite. Penman encoding is the most common method for AMR graph linearization, representing graphs as \textit{tree}-like structures. It uses variables (e.g. \texttt{s}, \texttt{s2} in Figure~\ref{fig:amr_penman_distance}) as node IDs to manage co-references. In addition, parentheses represent nested structures of AMR graphs. However, Penman has key limitations to training a seq2seq model:  

1)~\textbf{Parent-Child Distance}: Parent and child nodes may appear far apart in the linearized text, despite being closely connected in the graph. For example, in Figure~\ref{fig:amr_penman_distance}, \texttt{seem-01} and \texttt{we} are encoded distant in Penman format (highlighted in \textcolor{red}{red}) despite their proximity in the graph of Figure \ref{fig:amr_graph_distance}. This is observed when a preceding sibling node has a deep sub-graph. We hypothesize that this long distance increases the difficulty of learning strong parent-child connections, especially in deeper graphs. 
2)~\textbf{Inverse Roles}: 
Penman represents a graph in a tree-based format. To be specific, when a node has multiple parent nodes (node re-entrancy), the child node is duplicated to maintain a single-rooted tree structure. To fit an AMR graph into a tree structure, Penman introduces inverse roles by rewriting \texttt{:relation} as \texttt{:relation-of} (see Figure \ref{fig:amr_penman_distance} where inverse roles are highlighted in \textcolor{blue}{blue}). This increases the number of relations the model must learn, potentially complicating training and reducing model performance. Figure~\ref{fig:amr_without_invroles} shows how inverse roles are unnecessary in a graph-based representation.

To address these issues, we propose an alternative triple-based format for AMR graph linearization. A triple consists of a parent node, a child node, and a relation type between them, ensuring that parent and child nodes remain adjacent in the linearized text. This format also eliminates inverse roles by replacing \texttt{(node A, relation-of, node B)} with \texttt{(node B, relation, node A)}. In the rest of the paper, we compare Penman and triple-based formats with examples, highlighting their strengths and limitations in training a seq2seq AMR parser. Our contributions to seq2seq AMR parsing are:
\begin{itemize}[noitemsep]
    \item A triple-based linearization method for training seq2seq AMR parsers.  
    \item A detailed comparison with Penman linearization, focusing on performance across varying graph depths and lengths, and identifying areas for improvement.  
\end{itemize}

\section{Related Work}
Triple encoding has been used in relation extraction \citep{huguet-cabot-navigli-2021-rebel-relation,ye2021contrastive,saxena-etal-2022-sequence} and discourse representation structure (DRS) parsing \citep{van-noord-etal-2018-evaluating,van-noord-etal-2018-exploring}. The closest to our approach is \citet{van-noord-etal-2018-evaluating}, who convert AMR graphs into DRS triples. However, their representation differs from ours by mapping AMR relations to DRS roles and adding extra information that does not exist in AMR. In addition, they did not include training an AMR parser in their work. While triple encoding is widely used in relation extraction and DRS parsing, it has not been used for seq2seq AMR parsers. In our work, we propose using it to linearize AMR graphs, analyzing its strengths and weaknesses as an encoding method. 

\section{Methodology: Triple Linearization}
Triple representation mitigates the challenges of Penman linearization (described in Section \ref{sec:introduction}) by encoding a graph as a set of triples. Figure~\ref{fig:amr_triple} illustrates that the parent-child nodes, which were distantly located in Figure~\ref{fig:amr_penman_distance}, are now adjacent in the triple representation (highlighted in \textcolor{red}{red}). We hypothesize this helps the model learn direct parent-child relationships, especially in deeper graphs. Triples also eliminate inverse roles by reversing the order of two nodes, as shown in Figure~\ref{fig:amr_triple} (highlighted in \textcolor{blue}{blue}).

\begin{figure}
    \centering
    \begin{tiny}
    \begin{Verbatim}[commandchars=\\\{\}]
s instance seem-01          s2 instance see-01  p instance person   
a instance any              n instance nutter   d instance dig-01   
a2 instance acknowledge-01  t instance thing    t2 instance true-01 
i instance it               w instance we       e instance ever     
s polarity -                s ARG1 s2           s2 ARG0 w           
s2 ARG1 p                   p mod a             p mod n             
d \textbf{\textcolor{blue}{ARG0}} p                    a2 \textbf{\textcolor{blue}{ARG0}} p           a2 ARG1 t           
t2 \textbf{\textcolor{blue}{ARG1}} t                   t2 location i       \textbf{\textcolor{red}{s}} ARG2 \textbf{\textcolor{red}{w}}            
s time e
    \end{Verbatim}
    \end{tiny}
    \caption{Triple linearization of the graph in Figure \ref{fig:amr_penman_distance}.}
    \label{fig:amr_triple}
\end{figure}

The triple format, widely used for graph representation (e.g., RDF), aligns better with AMR’s graph structure than Penman’s tree-based encoding. Its flexibility supports graphs with multiple roots or re-entrancies, making it potentially suitable for broader semantic frameworks. To assess its utility, we trained seq2seq AMR parsers using triple and Penman formats.  

Despite its advantages, triple linearization can result in verbose linearization, slowing down the learning process, and may be less effective at capturing nested structures of graphs compared to Penman. This study evaluates both linearization methods to train seq2seq AMR parsers, exploring: \textbf{(1)} whether triple representation improves AMR parsing; \textbf{(2)} which graphs benefit most from triple representation, such as those with deep structures or large size, and \textbf{(3)} if combining triple and Penman representations enhances parsing performance. 

Experiments involve training models respectively with triple, Penman, and both formats (multi-task learning). Using both formats may serve as a form of data augmentation, as it effectively doubles the training data by representing one example in two linearized formats. We train and evaluate our model with English AMR~3.0 \citep{knight-etal-2020-ldc2020} data. We evaluate our model using \textsc{smatch} \citep{cai-knight-2013-smatch} score by counting the matching triples between two graphs. We analyze results by graph depth and size to determine which types of graphs benefit from different encoding methods. 

\paragraph{Triple linearization strategies}
To linearize AMR graphs in triples, we extract all triples and unfold them in depth-first search order using the PENMAN library.\footnote{\url{https://penman.readthedocs.io/en/}, version 1.3.0} Four linearization strategies are applied, varying in whether variables\footnote{Removing variables is a common pre-processing strategy for seq2seq AMR parsing \citep{konstas-etal-2017-neural,vannoord-et-bos-2017-neural}. This leads to information loss but effectively reduces data sparsity for training.} or inverse roles are retained. We provide an example for each linearization type in Table~\ref{tab:triple_linearization} and Figure~\ref{fig:amr_graph_for_linearization_examples} of Appendix~\ref{sec:appendix}. Each model is named based on linearization type as follows:
\begin{itemize}[noitemsep]
    \item \textbf{Triple\_X\_var\_X\_invrole}: Variables and inverse roles are removed. Variables are replaced by node names, and inverse roles are converted by reversing node order.\footnote{For the models discussed in this article, variables and inverse roles are removed in this manner.} Reversing inverse roles reduces the number of relation types from 155 to 115 in our training data. Triples are separated by a pipe symbol (|). 
    \item \textbf{Triple\_X\_var\_O\_invrole}: Variables are removed, but inverse roles are retained.
    \item \textbf{Triple\_O\_var\_O\_invrole}: Both variables and inverse roles are retained. Variables and their instances are represented as triples with the \texttt{instance} relation (e.g., \texttt{f instance fruit}). This approach is the most comprehensive, as no information is lost from the original graph during linearization.
    \item \textbf{Triple\_O\_var\_X\_invrole}: Variables are retained, but inverse roles are removed.
\end{itemize}

\section{Experimental Setup}
Models are trained using the large mBART model \citep{tang-etal-2021-multilingual} on each linearization type.\footnote{The model was chosen based on our goal of developing a multilingual system, which was not covered in this article.} 
\subsection{Baseline}
To compare our method with existing approaches using Penman encoding, we trained a model on AMR graphs linearized using Penman encoding, which serves as our baseline. Note that maintaining inverse roles is a necessary aspect of Penman encoding and \textbf{X\_invrole} types are not available for Penman encoding. For training, we employed the same mBART model and trained two models as follows (see Table~\ref{tab:penman_linearization} for examples):
\begin{itemize}[noitemsep]
    \item\textbf{Penman\_X\_var\_O\_invrole}: Variables are removed, but inverse roles are retained.\footnote{We used the script from \citet{vannoord-et-bos-2017-neural}.}
    \item \textbf{Penman\_O\_var\_O\_invrole}: Both variables and inverse roles are retained.
\end{itemize}

\subsection{Multi-task Learning}

As mentioned in Section~\ref{sec:introduction}, combining triple and Penman encodings may offer complementary benefits. To test this, we trained models in a multi-task learning framework by merging two differently encoded datasets. During training, the model learns from shuffled examples with a token indicating the encoding type. For predictions,  the model is instructed to use either triple or Penman encoding to use the corresponding decoding strategy to reconstruct graphs. We trained four models:

\begin{itemize}[noitemsep]
\item \textbf{Multi\_tri\_O\_var\_O\_invrole}: Both variables and inverse roles are retained. The main task is triple encoding, with Penman encoding as an auxiliary task. This means that the best model is selected based on the performance on the validation set using triple encoding, while Penman encoding is treated as an auxiliary task to help the triple learning. 
\item \textbf{Multi\_penman\_O\_var\_O\_invrole}: Both variables and inverse roles are retained. The main task is Penman encoding, with triple encoding as an auxiliary. 
\item \textbf{Multi\_tri\_X\_var\_O\_invrole}: Variables are removed but inverse roles are retained. The best model is chosen based on the model's performance in triple prediction.
\item \textbf{Multi\_penman\_X\_var\_O\_invrole}: Variables are removed but inverse roles are retained. The best model is selected based on performance in Penman prediction.
\end{itemize}

\section{Results and Insights}

\begin{table}
\centering
\resizebox{\linewidth}{!}{%
\begin{tabular}{@{}lll@{}}
\toprule
Model                             & \textit{The Little Prince} & AMR 3.0 \\ \midrule
Triple\_O\_var\_O\_invrole        & $ 76.2 \pm 0.3$ & $80.0 \pm 0.2 $ \\
Triple\_O\_var\_X\_invrole        & $ 76.1 \pm 0.5$ & $80.3 \pm 0.1 $ \\
Triple\_X\_var\_O\_invrole        & $ 76.5 \pm 0.2$ & $78.7 \pm 0.1 $ \\
Triple\_X\_var\_X\_invrole        & $ 76.2 \pm 0.2$ & $78.9 \pm 0.8 $ \\ \midrule
Penman\_O\_var\_O\_invrole        & $ \textbf{77.0} \pm 0.4$ & $\textbf{80.9} \pm 0.2 $ \\
Penman\_X\_var\_O\_invrole        & $ 76.7 \pm 0.1$ & $80.2 \pm 0.1 $ \\ \midrule
Multi\_tri\_O\_var\_O\_invrole    & $ \underline{76.9} \pm 0.2$ & $80.3 \pm 0.1 $ \\
Multi\_tri\_X\_var\_O\_invrole    & $ 76.3 \pm 0.2$ & $78.9 \pm 0.1 $ \\
Multi\_penman\_O\_O\_invrole      & $ 76.7 \pm 0.2$ & $ \underline{80.6} \pm 0.3 $ \\
Multi\_penman\_X\_var\_O\_invrole & $ 76.1 \pm 0.1$ & $79.8 \pm 0.1 $ \\ \bottomrule
\end{tabular}}
\caption{\textsc{smatch} scores for evaluation (with the highest scores in \textbf{bold} and the second-highest scores \underline{underlined}).}
\label{tab:eng_smatch}
\end{table}

\paragraph{Global results.} 

Table \ref{tab:eng_smatch} presents results on two test sets: \textit{The Little Prince}\footnote{\url{https://github.com/flipz357/AMR-World}} and AMR~3.0. The Penman single-task model with variables (\textbf{Penman\_O\_var\_O\_invrole}) performs best on both test sets (77.0 and 80.9, respectively). Among single-task triple models, \textbf{Tri\_O\_var\_O\_invrole} achieves the best results, with a marginal gap from the best model ($\leq$1 \textsc{smatch}).  

Preserving variables consistently improves performance, contradicting the assumption that removing them aids learning by reducing data sparsity. This suggests variable removal leads to critical information loss. In addition, learning from Penman encoding while performing an auxiliary triple task reduces performance, whereas the reverse improves it. This indicates Penman encoding provides structural information beneficial to triple encoding but not vice versa.  

Within triple linearization, removing inverse roles improves performance on AMR~3.0 but not \textit{The Little Prince}. Given that AMR~3.0 includes longer sentences, removing inverse roles may benefit longer sentences while harming shorter ones. However, the inconsistency across test sets could also suggest inverse roles have only a marginal effect.

\paragraph{Triple linearization does not improve learning for deeper graphs.}

We hypothesized that triple linearization could enhance training by positioning child and parent nodes closer, especially when a preceding sibling has a deep subgraph. In Penman encoding, such cases place these nodes farther apart. Assuming this issue is more common in deeper graphs, we analyzed results by reference graph depth (distance from the root to the furthest node), focusing on AMR 3.0, which has greater depth variety than \textit{The Little Prince}.  

Our results (Figure \ref{fig:smatch_per_depth}, Appendix) show \textsc{smatch} scores by depth align with overall scores, indicating no learning benefit for deeper graphs with triple encoding. The best model, \textbf{Penman\_O\_var\_O\_inverserole}, consistently performed best across depths. This contradicts our hypothesis that triple encoding benefits seq2seq AMR learning by bringing parent-child nodes closer together. Instead, the results emphasize the benefit of Penman’s concise graph representation and its ability to explicitly encode nested structures, which play a more critical role in model performance.

\paragraph{Triple linearization does not improve learning for longer graphs.} 

We also analyzed performance by graph length (i.e., token count in the linearized reference graph), assuming verbose triple encoding would degrade performance on longer graphs. Since graph depth and length are not always correlated, results may differ from the depth analysis.  Figure~\ref{fig:smatch_per_length} in the Appendix shows the results per graph length with token counts of reference graphs grouped into buckets of 50 for clarity. For shorter graphs, the gap between models is smaller, but as the graphs become longer, the performance of triple models decreases more noticeably, resulting in a wider gap. This supports our earlier hypothesis regarding the limitations of triple encoding: its verbose representation likely contributes to the observed performance degradation on lengthy graphs.
\section{Conclusion}

We introduced triple linearization as an alternative to Penman linearization, hypothesizing that it could improve training for several reasons: (1) parent and child nodes are always located together in a triple, (2) the elimination of inverse roles may simplify training by reducing the number of relations, (3) and triples more closely resemble the underlying graph structure, while Penman encoding represents a graph in a more tree-like format. Contrary to our hypothesis, Penman has proven to be a more effective linearization method to train a seq2seq parsing. However, the gap between the best Penman model and certain triple-based models is marginal. Our results show a potential to train a seq2seq AMR parser that predicts a graph directly (not a tree-based representation) while maintaining equivalent performance. Notably, the model's output in triples more naturally aligns with AMR's graph structure than Penman. Our code to train and evaluate the model is available on \url{https://github.com/Emvista/Triple_AMR_Parser.git}. 

\section*{Acknowledgments}

We thank reviewers for their helpful comments and Cécile Macaire for proof-reading this article. Jeongwoo Kang and Maximin Coavoux gratefully acknowledge the support of the French National Research Agency (grant ANR-23-CE23-0017-01). This work was granted access to the HPC resources of IDRIS under the allocation 2024-AD011012853R2 made by GENCI.

\bibliography{custom}

\begin{thebibliography}{15}
\providecommand{\natexlab}[1]{#1}

\bibitem[{Banarescu et~al.(2013)Banarescu, Bonial, Cai, Georgescu, Griffitt,
  Hermjakob, Knight, Koehn, Palmer, and
  Schneider}]{banarescu-etal-2013-abstract}
Laura Banarescu, Claire Bonial, Shu Cai, Madalina Georgescu, Kira Griffitt, Ulf
  Hermjakob, Kevin Knight, Philipp Koehn, Martha Palmer, and Nathan Schneider.
  2013.
\newblock \href {https://aclanthology.org/W13-2322} {{A}bstract {M}eaning
  {R}epresentation for sembanking}.
\newblock In \emph{Proceedings of the 7th Linguistic Annotation Workshop and
  Interoperability with Discourse}, pages 178--186, Sofia, Bulgaria.
  Association for Computational Linguistics.

\bibitem[{Bonial et~al.(2023)Bonial, Foresta, Fung, Hayes, Osteen, Arkin,
  Hedegaard, and Howard}]{bonial-etal-2023-abstract}
Claire Bonial, Julie Foresta, Nicholas~C. Fung, Cory~J. Hayes, Philip Osteen,
  Jacob Arkin, Benned Hedegaard, and Thomas Howard. 2023.
\newblock \href {https://aclanthology.org/2023.dmr-1.4} {{A}bstract {M}eaning
  {R}epresentation for grounded human-robot communication}.
\newblock In \emph{Proceedings of the Fourth International Workshop on
  Designing Meaning Representations}, pages 34--44, Nancy, France. Association
  for Computational Linguistics.

\bibitem[{Bonial et~al.(2019)Bonial, Donatelli, Ervin, and
  Voss}]{bonial-etal-2019-abstract}
Claire~N. Bonial, Lucia Donatelli, Jessica Ervin, and Clare~R. Voss. 2019.
\newblock \href {https://doi.org/10.7275/v3c5-yd35} {{A}bstract {M}eaning
  {R}epresentation for human-robot dialogue}.
\newblock In \emph{Proceedings of the Society for Computation in Linguistics
  ({SC}i{L}) 2019}, pages 236--246.

\bibitem[{Cai and Knight(2013)}]{cai-knight-2013-smatch}
Shu Cai and Kevin Knight. 2013.
\newblock \href {https://aclanthology.org/P13-2131} {{S}match: an evaluation
  metric for semantic feature structures}.
\newblock In \emph{Proceedings of the 51st Annual Meeting of the Association
  for Computational Linguistics (Volume 2: Short Papers)}, pages 748--752,
  Sofia, Bulgaria. Association for Computational Linguistics.

\bibitem[{Huang et~al.(2023)Huang, Iyer, Hsu, Kumar, Chang, and
  Galstyan}]{huang-etal-2023-paraamr}
Kuan-Hao Huang, Varun Iyer, I-Hung Hsu, Anoop Kumar, Kai-Wei Chang, and Aram
  Galstyan. 2023.
\newblock \href {https://doi.org/10.18653/v1/2023.acl-long.447} {{P}ara{AMR}: A
  large-scale syntactically diverse paraphrase dataset by {AMR}
  back-translation}.
\newblock In \emph{Proceedings of the 61st Annual Meeting of the Association
  for Computational Linguistics (Volume 1: Long Papers)}, pages 8047--8061,
  Toronto, Canada. Association for Computational Linguistics.

\bibitem[{Huguet~Cabot and
  Navigli(2021)}]{huguet-cabot-navigli-2021-rebel-relation}
Pere-Llu{\'\i}s Huguet~Cabot and Roberto Navigli. 2021.
\newblock \href {https://doi.org/10.18653/v1/2021.findings-emnlp.204} {{REBEL}:
  Relation extraction by end-to-end language generation}.
\newblock In \emph{Findings of the Association for Computational Linguistics:
  EMNLP 2021}, pages 2370--2381, Punta Cana, Dominican Republic. Association
  for Computational Linguistics.

\bibitem[{Knight et~al.(2020)Knight, Badarau, Baranescu, Bonial, Bardocz,
  Griffitt, Hermjakob, Marcu, Palmer, O'Gorman, and
  Schneider}]{knight-etal-2020-ldc2020}
Kevin Knight, Bianca Badarau, Laura Baranescu, Claire Bonial, Madalina Bardocz,
  Kira Griffitt, Ulf Hermjakob, Daniel Marcu, Martha Palmer, Tim O'Gorman, and
  Nathan Schneider. 2020.
\newblock \href {https://doi.org/10.35111/44cy-bp51} {Abstract meaning
  representation (amr) annotation release 3.0 ldc2020t02}.
\newblock Philadelphia: Linguistic Data Consortium.

\bibitem[{Konstas et~al.(2017)Konstas, Iyer, Yatskar, Choi, and
  Zettlemoyer}]{konstas-etal-2017-neural}
Ioannis Konstas, Srinivasan Iyer, Mark Yatskar, Yejin Choi, and Luke
  Zettlemoyer. 2017.
\newblock \href {https://doi.org/10.18653/v1/P17-1014} {Neural {AMR}:
  Sequence-to-sequence models for parsing and generation}.
\newblock In \emph{Proceedings of the 55th Annual Meeting of the Association
  for Computational Linguistics (Volume 1: Long Papers)}, pages 146--157,
  Vancouver, Canada. Association for Computational Linguistics.

\bibitem[{Saxena et~al.(2022)Saxena, Kochsiek, and
  Gemulla}]{saxena-etal-2022-sequence}
Apoorv Saxena, Adrian Kochsiek, and Rainer Gemulla. 2022.
\newblock \href {https://doi.org/10.18653/v1/2022.acl-long.201}
  {Sequence-to-sequence knowledge graph completion and question answering}.
\newblock In \emph{Proceedings of the 60th Annual Meeting of the Association
  for Computational Linguistics (Volume 1: Long Papers)}, pages 2814--2828,
  Dublin, Ireland. Association for Computational Linguistics.

\bibitem[{Tang et~al.(2021)Tang, Tran, Li, Chen, Goyal, Chaudhary, Gu, and
  Fan}]{tang-etal-2021-multilingual}
Yuqing Tang, Chau Tran, Xian Li, Peng-Jen Chen, Naman Goyal, Vishrav Chaudhary,
  Jiatao Gu, and Angela Fan. 2021.
\newblock \href {https://doi.org/10.18653/v1/2021.findings-acl.304}
  {Multilingual translation from denoising pre-training}.
\newblock In \emph{Findings of the Association for Computational Linguistics:
  ACL-IJCNLP 2021}, pages 3450--3466, Online. Association for Computational
  Linguistics.

\bibitem[{van Noord et~al.(2018{\natexlab{a}})van Noord, Abzianidze, Haagsma,
  and Bos}]{van-noord-etal-2018-evaluating}
Rik van Noord, Lasha Abzianidze, Hessel Haagsma, and Johan Bos.
  2018{\natexlab{a}}.
\newblock \href {https://aclanthology.org/L18-1267} {Evaluating scoped meaning
  representations}.
\newblock In \emph{Proceedings of the Eleventh International Conference on
  Language Resources and Evaluation ({LREC} 2018)}, Miyazaki, Japan. European
  Language Resources Association (ELRA).

\bibitem[{van Noord et~al.(2018{\natexlab{b}})van Noord, Abzianidze, Toral, and
  Bos}]{van-noord-etal-2018-exploring}
Rik van Noord, Lasha Abzianidze, Antonio Toral, and Johan Bos.
  2018{\natexlab{b}}.
\newblock \href {https://doi.org/10.1162/tacl_a_00241} {Exploring neural
  methods for parsing discourse representation structures}.
\newblock \emph{Transactions of the Association for Computational Linguistics},
  6:619--633.

\bibitem[{van Noord and Bos(2017)}]{vannoord-et-bos-2017-neural}
Rik van Noord and Johan Bos. 2017.
\newblock \href {https://www.clinjournal.org/clinj/article/view/72} {Neural
  semantic parsing by character-based translation: Experiments with abstract
  meaning representations}.
\newblock \emph{Computational Linguistics in the Netherlands Journal},
  7:93–108.

\bibitem[{Wein and Schneider(2024)}]{wein-schneider-2024-lost}
Shira Wein and Nathan Schneider. 2024.
\newblock \href {https://aclanthology.org/2024.eacl-long.45} {Lost in
  translationese? reducing translation effect using {A}bstract {M}eaning
  {R}epresentation}.
\newblock In \emph{Proceedings of the 18th Conference of the European Chapter
  of the Association for Computational Linguistics (Volume 1: Long Papers)},
  pages 753--765, St. Julian{'}s, Malta. Association for Computational
  Linguistics.

\bibitem[{Ye et~al.(2021)Ye, Zhang, Deng, Chen, Tan, Huang, and
  Chen}]{ye2021contrastive}
Hongbin Ye, Ningyu Zhang, Shumin Deng, Mosha Chen, Chuanqi Tan, Fei Huang, and
  Huajun Chen. 2021.
\newblock Contrastive triple extraction with generative transformer.
\newblock In \emph{Proceedings of the AAAI conference on artificial
  intelligence}, volume~35, pages 14257--14265.

\end{thebibliography}
\onecolumn

\appendix

\section{Appendix}\label{sec:appendix}

\begin{figure}[h]
    \centering
    \includegraphics[width=0.75\linewidth]{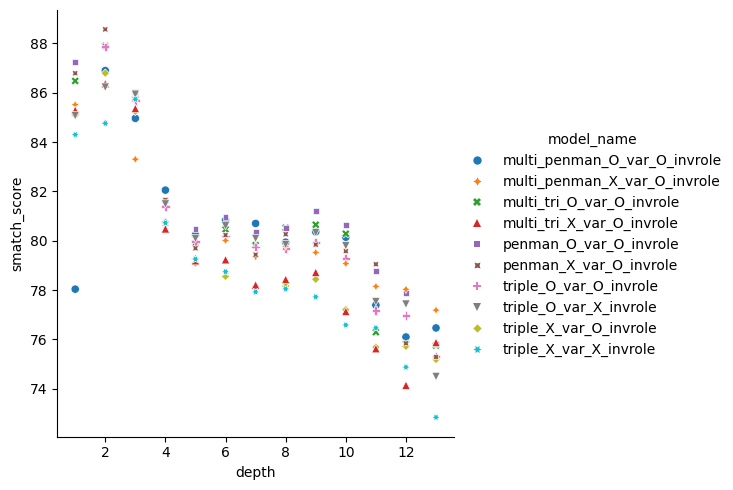}
    \caption{\textsc{smatch} score per graph depth.}
    \label{fig:smatch_per_depth}
\end{figure}

\begin{figure}[h]
    \centering
    \includegraphics[width=0.75\linewidth]{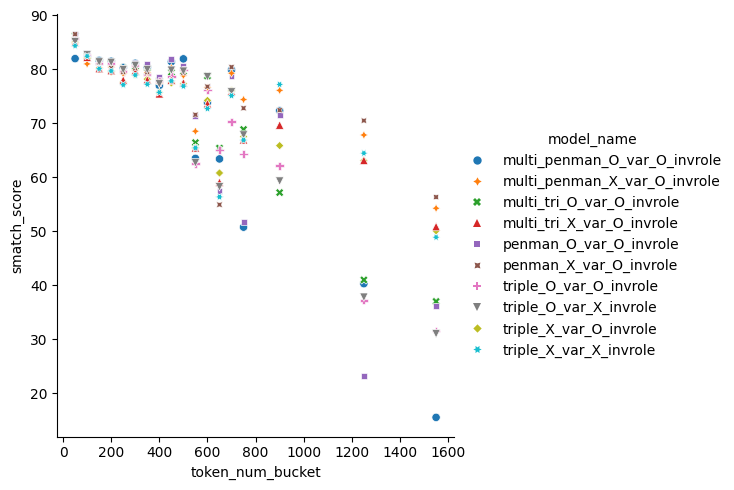}
    \caption{\textsc{smatch} score per graph length. The length is measured by the number of tokens in a linearized graph and token counts are grouped into buckets of 50 for clarity.}
    \label{fig:smatch_per_length}
\end{figure}

\begin{figure}
    \centering
    \includegraphics[width=0.45\linewidth]{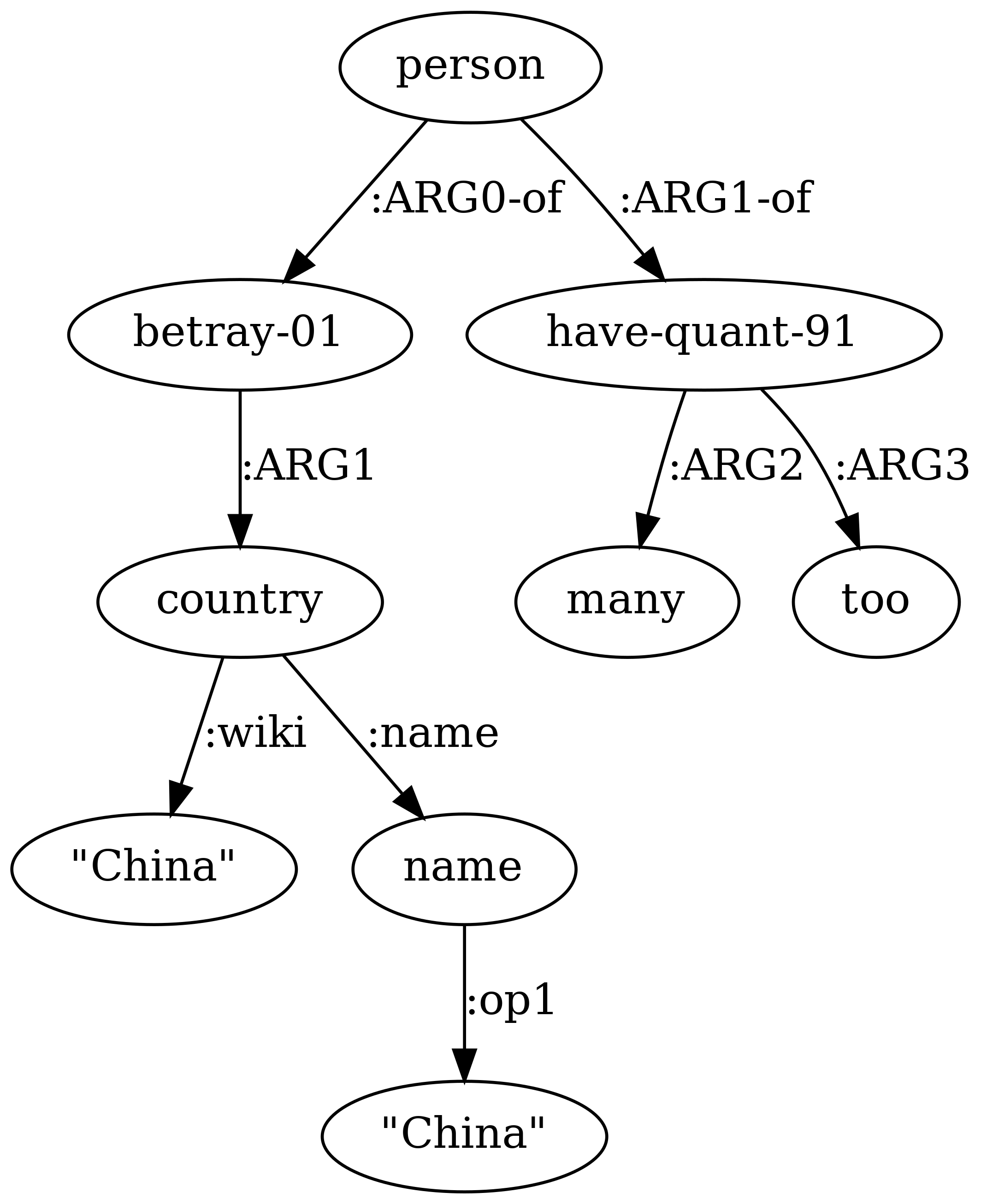}
    \caption{AMR graph for ``There are too many traitors of China!''. This example is drawn from AMR~3.0 dataset \citep{knight-etal-2020-ldc2020}.}
    \label{fig:amr_graph_for_linearization_examples}
\end{figure}

\begingroup
\setlength{\tabcolsep}{4pt} 
\renewcommand{\arraystretch}{1.5} 
\begin{table} \tiny
\resizebox{\linewidth}{!}{%
\begin{tabularx}
{1.01\linewidth}{|>{\hsize=.28\hsize}X|>{\hsize=.25\hsize}X|>{\hsize=.24\hsize}X|>{\hsize=.24\hsize}X|}
\hline
Triple\_X\_var\_X\_invrole & Triple\_X\_var\_O\_invrole & Triple\_O\_var\_O\_invrole & Triple\_O\_var\_X\_invrole \\ \hline
\ \newline \ \newline \ \newline \ \newline \ \newline \ \newline
\texttt{person ARG0-of betray-01 |\newline betray-01 ARG1 country |\newline country name " China " |\newline person ARG1-of have-quant-91 |\newline have-quant-91 ARG2 many |\newline have-quant-91 ARG3 too} & \ \newline \ \newline \ \newline \ \newline \ \newline \ \newline \texttt{betray-01 ARG0 person |\newline betray-01 ARG1 country |\newline country name " China " |\newline have-quant-91 ARG1 person |\newline have-quant-91 ARG2 many |\newline have-quant-91 ARG3 too} & \texttt{p instance person |\newline b instance betray-01 |\newline c instance country |\newline h instance have-quant-91 |\newline m instance many |\newline t instance too |\newline p ARG0-of b |\newline b ARG1 c |\newline c name " China " |\newline p ARG1-of h |\newline h ARG2 m |\newline h ARG3 t} & \texttt{p instance person |\newline b instance betray-01 |\newline c instance country |\newline h instance have-quant-91 |\newline m instance many |\newline t instance too |\newline b ARG0 p |\newline b ARG1 c |\newline c name " China " |\newline h ARG1 p |\newline h ARG2 m |\newline h ARG3 t} \\ \hline
\end{tabularx}}
\caption[Examples for different AMR linearization with triple encoding.]{Triple encoding examples of Figure \ref{fig:amr_graph_for_linearization_examples}.}
\label{tab:triple_linearization}
\end{table}
\endgroup

\begingroup
\setlength{\tabcolsep}{4pt} 
\renewcommand{\arraystretch}{1.5} 
\begin{table}
\scriptsize
\begin{tabularx}{1.01\linewidth}{|X|X|}
\hline
Penman\_X\_var\_O\_invrole & Penman\_O\_var\_O\_invrole \\ \hline \texttt{( person :ARG0-of ( betray-01 :ARG1 ( country :name " China " ) ) :ARG1-of ( have-quant-91 :ARG2 ( many ) :ARG3 ( too ) ) )} & 
 \texttt{( p / person :ARG0-of ( b / betray-01 :ARG1 ( c / country :name " China " ) :ARG1-of ( h / have-quant-91 :ARG2 ( m / many ) :ARG3 ( t / too ) ) )} \\ \hline
\end{tabularx}
\caption[Examples for different AMR linearization with Penman endoing.]{Penman encoding examples of Figure \ref{fig:amr_graph_for_linearization_examples}.}
\label{tab:penman_linearization}
\end{table}
\endgroup

\end{document}